\documentclass[11pt]{article}

\usepackage[margin=1in]{geometry}
\usepackage[T1]{fontenc}
\usepackage[utf8]{inputenc}
\usepackage{lmodern}
\usepackage{amsmath,amssymb}
\usepackage{booktabs}
\usepackage{array}
\usepackage{graphicx}
\usepackage{hyperref}
\usepackage{xcolor}

\hypersetup{
  colorlinks=true,
  linkcolor=blue!60!black,
  citecolor=blue!60!black,
  urlcolor=blue!60!black
}

\title{FiLM-Coordinated Dual-Branch Transformer for Global-Local Dependency Modeling in Language Modeling}
\author{Zhiqiang Zhou \and Xu Ling \and Junliang Dai}
\date{}

\begin{document}

\maketitle

\begin{abstract}
Standard Transformers rely on a single self-attention pathway to model both global dependencies and local patterns, which creates an inherent tension between long-range structural reasoning and fine-grained local representation learning. To address this issue, we propose a FiLM-coordinated dual-branch Transformer for language modeling. The architecture explicitly separates a global branch and a local branch within each layer, and uses feature-wise linear modulation (FiLM) to enable dynamic cross-branch coordination instead of simple concatenation or static addition.

The key intuition is that the two branches process different dependency views of the same input sequence, making channel-wise calibration more appropriate than heavy token-level interaction. Based on this view, we design a bidirectional FiLM module in which each branch generates per-channel scaling and shifting parameters to condition the representation of the other branch. This enables input-dependent coordination within the layer. We further analyze the design through directional ablations, local-window ablations, and FiLM-generator capacity ablations, and examine the learned modulation patterns from the perspectives of channel selectivity, layerwise variation, and input adaptivity.

Experiments on multiple small-scale language modeling settings show that the proposed structure yields stable gains over same-width single-branch baselines and weakened dual-branch variants under a fixed lightweight configuration. On both TinyShakespeare and a 1M-character subset of WikiText-2, the full dual-branch FiLM model outperforms same-width single-branch and dual-concatenation baselines. Multi-seed results further indicate that the gains are not due to random initialization. Mechanistic analyses suggest that FiLM learns non-uniform, input-dependent, and layer-dependent modulation patterns rather than static scaling.

At the same time, parameter-matched fairness baselines with widened single-branch models show that the current design still has room for improvement in parameter efficiency. Therefore, the main contribution of this paper is not a stronger fusion head, but a new FiLM-coordinated dual-branch Transformer structure that more effectively integrates global and local dependencies under a fixed lightweight budget.
\end{abstract}

\section{Introduction}

Attention is at the core of modern large language models, yet standard self-attention typically uses a single pathway to handle both long-range dependency modeling and local pattern extraction. As models and contexts scale up, this unified treatment starts to show an obvious tension: global modeling demands broader receptive fields and stronger structural abstraction, while local pattern recognition benefits from stable neighborhood inductive bias. These two objectives are not incompatible, but they are not necessarily served efficiently by the same attention pathway.

A natural alternative is to build a dual-branch Transformer in which one branch focuses on global dependencies and the other emphasizes local context, making this division of labor explicit within each layer. Related hybrid ideas have appeared at the inter-layer level. For example, Jamba~\cite{jamba2024,jamba152024} alternates different computational units across layers to improve the efficiency-performance trade-off. However, it remains unclear how two same-layer branches derived from the same input should coordinate in a principled way. Merely placing two branches in parallel is not sufficient; the key question is how they exchange and calibrate information.

Existing work often reduces this issue to a comparison among fusion operators such as concatenation, weighted summation, or cross-attention. This framing tends to shift the focus toward which fusion head performs best, while the more fundamental question is how to design an effective coordination mechanism for a dual-branch structure. In our setting, the global and local branches are not independent information sources. Instead, they are two dependency views of the same sequence. This makes dynamic channel-wise calibration more suitable than heavy retrieval-style interaction.

Motivated by this observation, we propose a FiLM-coordinated dual-branch Transformer. Bidirectional FiLM modules generate per-channel scaling and shifting parameters across the two branches so that each branch can condition the feature expression of the other based on its own state. In this design, FiLM is not treated as an interchangeable fusion trick. It becomes the core coordination mechanism inside the dual-branch structure: the global branch captures long-range structure, the local branch captures fine-grained contextual patterns, and FiLM aligns the two views in an input-dependent manner.

Experiments show that this structure-oriented perspective explains the observed results better than a simple fusion comparison. The full dual-branch FiLM structure outperforms single-branch baselines and weakened dual-branch variants under a fixed lightweight setting. Bidirectional modulation performs better than one-way modulation. Window-size and generator-capacity ablations suggest that the gains do not rely on extreme hyperparameters. Multi-seed and cross-dataset results further support the stability of the effect. Mechanistic analyses also indicate that the learned modulation is channel-selective, layer-dependent, and input-adaptive rather than uniform.

Therefore, the central contribution of this paper is not to prove that FiLM is the best fusion mechanism in general, but to formulate and validate a FiLM-coordinated dual-branch Transformer as a new structural design for joint global-local dependency modeling within a layer.

Our contributions are summarized as follows:
\begin{enumerate}
  \item We propose a FiLM-coordinated dual-branch Transformer that separately models global and local dependencies within the same layer and uses bidirectional FiLM for dynamic cross-branch coordination.
  \item We reposition FiLM from a generic conditioning tool to a structural coordination module, arguing that in same-source dual-branch settings the relation between branches is better viewed as view calibration rather than feature concatenation.
  \item We validate the structure through main comparisons, multi-seed stability tests, cross-dataset evaluations, and structural ablations covering directional modulation, local-window size, FiLM-generator capacity, and branch-level baselines.
  \item We show through mechanistic analysis and parameter-matched fairness baselines that the gains are most evident under fixed lightweight budgets, while the current design still leaves room for improvement in parameter efficiency.
\end{enumerate}

\section{Related Work}

\subsection{Efficient Attention Variants}

The quadratic complexity of standard multi-head attention has motivated many efficient attention variants. Sparse attention methods reduce computation through block routing or fixed sparse patterns, with representative examples including MoBA~\cite{moba2025}, Native Sparse Attention~\cite{native_sparse2025}, Sparse Transformer~\cite{sparse_transformer2019}, and Longformer~\cite{longformer2020}. Local attention introduces window masks to strengthen neighborhood inductive bias, as exemplified by Mistral~\cite{mistral2023}. Compression-oriented attention such as MLA in DeepSeek-V2~\cite{deepseekv2_2024} and grouped-query attention~\cite{gqa2023} focus more on reducing KV-cache and inference overhead. Most of these methods optimize a single attention pathway rather than asking how different dependency biases should coordinate within the same layer.

\subsection{Hybrid and Dual-Path Structures}

Hybrid architectures have proven to be a practical route to improving modeling capacity. Jamba~\cite{jamba2024,jamba152024} alternates different computational blocks across layers, while mixture-of-experts models~\cite{switch2022,mixtral2024} route tokens across experts in the feed-forward stage. However, these approaches either operate across layers or rely on expert routing rather than addressing how two attention pathways inside the same layer should coordinate. Our goal is not to introduce another fusion head, but to propose a more suitable intra-layer coordination paradigm for a dual-path Transformer.

\subsection{Feature-Wise Modulation}

FiLM was originally introduced for conditional visual reasoning~\cite{film2018} and has since been widely used in style transfer~\cite{adain2017}, generative models~\cite{stylegan2019}, and controllable diffusion~\cite{controlnet2023}. Its core strength lies in fine-grained feature conditioning through per-channel affine transformation. This makes it well suited to calibrating multiple expressions within a shared semantic space. Our use of FiLM differs from prior settings in that FiLM is not an external conditioning interface; it is embedded inside the dual-branch Transformer as the structural coordination mechanism between the global and local branches.

\section{Method}

\subsection{Problem Formulation}

We study not the problem of selecting the best fusion head for two attention outputs, but the problem of designing a dual-branch Transformer structure that can stably coordinate a global branch and a local branch. Let the input sequence be $x$, and let branches $A$ and $B$ produce outputs $h_A=f_A(x)$ and $h_B=f_B(x)$, respectively. The coordination problem can then be written as finding an intra-layer coordination function $g(h_A,h_B)$ such that the two branches preserve their inductive biases while forming a stronger joint representation.

For analysis, we group common choices into three coordination paradigms. The first is concatenation, where the two branch outputs are combined and projected. The second is interaction, where cross-attention allows one branch to retrieve information from the other. The third is modulation, where one branch generates feature-wise modulation parameters for the other. Our central hypothesis is that in same-source dual-branch settings, the modulation paradigm is a better structural match than concatenation or heavy interaction.

\subsection{Why FiLM?}

We view the global and local branches as two dependency views of the same input rather than two unrelated sources of information. The global branch is better suited for long-range structural relationships, while the local branch is better for neighborhood patterns and short-range repetition. Under this view, the more appropriate operation is not to merge them with a static operator, but to let one view dynamically calibrate the reliability of the other.

FiLM provides exactly this kind of channel-wise calibration. Given a branch representation $h$, a lightweight MLP generates $\gamma$ and $\beta$ and applies
\begin{equation}
g(h) = (1+\gamma)\cdot h + \beta .
\end{equation}
Here $\gamma$ controls scaling and $\beta$ provides offset correction. Because $\gamma$ and $\beta$ are generated from input-dependent branch states, the modulation is inherently dynamic and conditional. In contrast, concatenation followed by a linear layer is effectively a static mixing operator, while cross-attention introduces heavier token-level computation between two already related views.

\subsection{FiLM-Coordinated Dual-Branch Structure}

The proposed architecture contains three components: a global branch, a local branch, and a bidirectional FiLM coordination module. The input is first encoded by the two branches in parallel. Each branch then generates modulation parameters for the other branch. The modulated outputs are added together and passed through a residual connection.

The data flow can be written as:
\begin{enumerate}
  \item The input $x$ is sent to the global branch to produce $h_A$.
  \item The input $x$ is sent to the local branch to produce $h_B$.
  \item $h_A$ generates $\gamma_A,\beta_A$ to modulate $h_B$.
  \item $h_B$ generates $\gamma_B,\beta_B$ to modulate $h_A$.
  \item We obtain $h'_A=(1+\gamma_B)\cdot h_A+\beta_B$.
  \item We obtain $h'_B=(1+\gamma_A)\cdot h_B+\beta_A$.
  \item The final output is $\mathrm{LayerNorm}(x+h'_A+h'_B)$.
\end{enumerate}

\subsection{Global and Local Branches}

In the current prototype, the global branch is responsible for broader information aggregation, while the local branch uses sliding-window attention to encode nearby context. The purpose is not to fully replicate industrial efficient-attention systems such as MoBA, but to construct a controlled and budget-aware prototype that is sufficient to test the structural hypothesis.

\subsection{FiLM Generator and Initialization}

The FiLM generator uses a lightweight MLP with zero initialization. This choice has two benefits. First, the structure starts from a weakly coordinated regime in early training, which avoids destructive random modulation. Second, it makes the learned modulation easier to interpret as progressively acquired coordination rather than as an accidental effect of initialization noise.

\begin{figure}[t]
  \centering
  \includegraphics[width=0.98\linewidth]{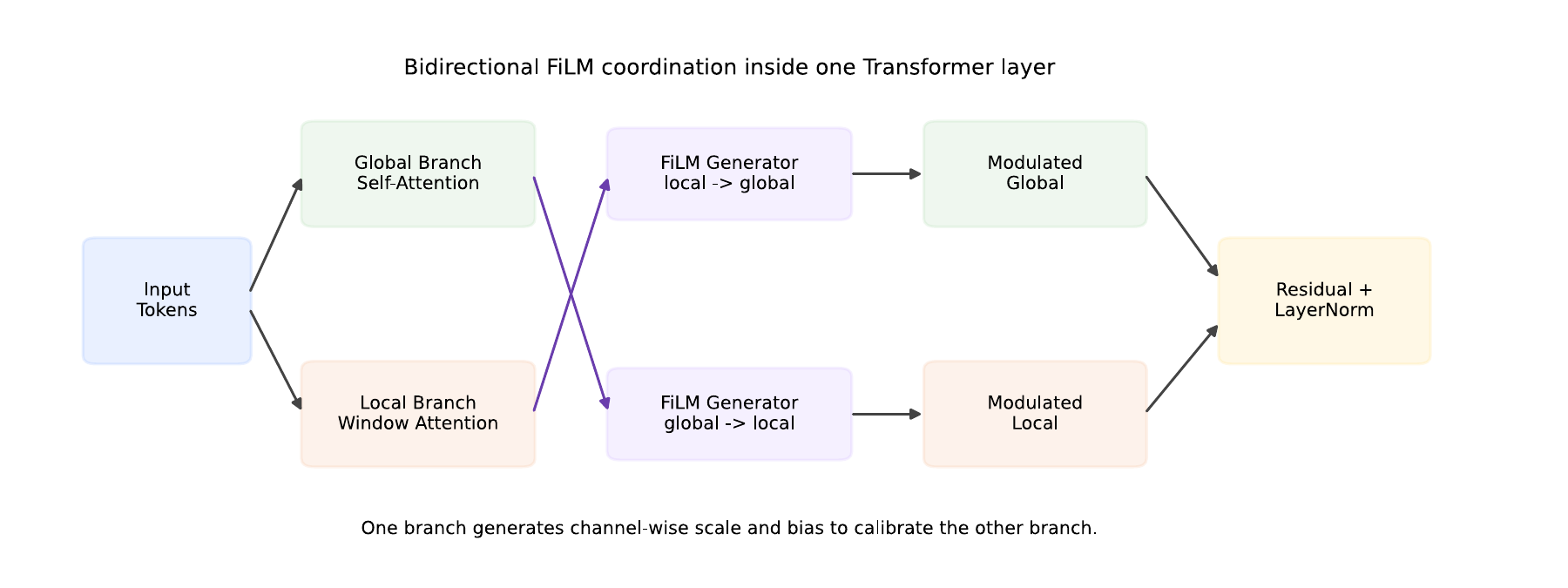}
  \caption{Single-layer overview of the FiLM-coordinated dual-branch Transformer. The global and local branches first model different dependency patterns in parallel, then generate per-channel scaling and shifting parameters for each other, and finally produce the layer output through residual integration.}
  \label{fig:architecture}
\end{figure}

Figure~\ref{fig:architecture} illustrates the overall structure. Unlike static fusion after simple parallel encoding, our method first encodes two dependency views and then uses bidirectional FiLM for dynamic channel-wise calibration before residual integration.

\section{Experiments}

\subsection{Experimental Setup}

Our experimental goal is a small-scale but closed-loop structural validation. We aim to answer three questions: (1) whether the FiLM-coordinated dual-branch structure improves over single-branch or weakened dual-branch baselines, (2) whether the gains remain stable across seeds and text settings, and (3) what role FiLM plays inside the structure.

\subsubsection{Model Configuration}

All experiments are built on the same lightweight decoder-only dual-branch Transformer scaffold. Unless otherwise specified, we use 4 layers, hidden size 192, 6 attention heads, sequence length 128, batch size 16, and 3 training epochs. The global branch models broader dependencies, and the local branch uses window attention with default window size 64. The FiLM generator uses a default hidden ratio of 2.0 with zero initialization.

\subsubsection{Data Settings}

We use two real-text settings and one controlled auxiliary setting. The real-text settings are TinyShakespeare and a fixed 1M-character subset of the WikiText-2 training split. The latter serves as a second validation setting with a clearly different text style while keeping the training budget similar to TinyShakespeare. The controlled auxiliary setting is used mainly for mechanism analysis rather than headline results.

\subsubsection{Baselines and Variants}

To distinguish structural value from coordination-module value, we use a two-level comparison. The first level is structural comparison among \texttt{single\_global}, \texttt{single\_local}, \texttt{dual\_concat}, and \texttt{dual\_film}. This level answers whether both the dual-branch structure and the full coordination design are necessary. The second level is coordination comparison among Add, Gate, Cross, and FiLM under the same dual-branch skeleton. This level is intended to clarify why FiLM is a better internal coordination mechanism, not to redefine the paper as a generic fusion benchmark.

\subsubsection{Metrics}

The main metrics are validation loss and perplexity. To analyze structural cost, we additionally report inference latency, peak memory, and tokens per second when available. For key experiments, we aggregate results over three random seeds (42/43/44). Mechanistic analyses further report FiLM statistics such as \texttt{mean\_abs\_gamma}, \texttt{var\_gamma}, and \texttt{topk\_gamma\_ratio}.

\subsection{Main Results: Is the FiLM-Coordinated Dual-Branch Structure Effective?}

Table~\ref{tab:tiny-structure} presents the main TinyShakespeare results.

\begin{table}[t]
  \centering
  \caption{Main TinyShakespeare results for single-branch, weakened dual-branch, and full dual-branch structures.}
  \label{tab:tiny-structure}
  \begin{tabular}{lcc}
    \toprule
    Variant & final\_val\_loss & final\_val\_ppl \\
    \midrule
    single\_global & 1.8231 & 6.1909 \\
    single\_local & 1.8193 & 6.1667 \\
    dual\_concat & 1.7886 & 5.9822 \\
    dual\_film & 1.6894 & 5.4117 \\
    \bottomrule
  \end{tabular}
\end{table}

\subsubsection{Structural Main Result}

The first question is whether the full dual-branch FiLM structure yields a genuine structural gain. On TinyShakespeare, \texttt{dual\_film} achieves the best result with \texttt{final\_val\_loss}=1.6894, outperforming \texttt{dual\_concat} at 1.7886 as well as both single-branch baselines. This suggests that the benefit does not come merely from adding another branch, but from the combination of dual-path structure and dynamic coordination.

\begin{figure}[t]
  \centering
  \includegraphics[width=0.8\linewidth]{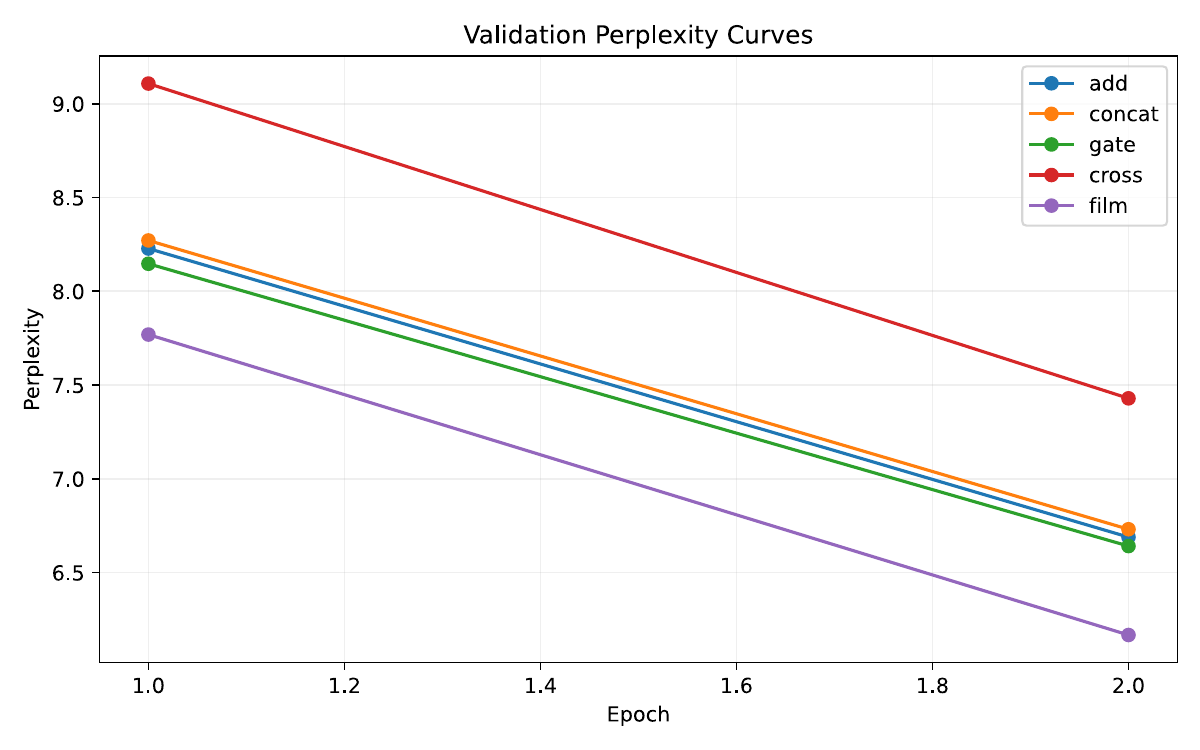}
  \caption{Validation perplexity curves on TinyShakespeare. The full dual-branch FiLM structure maintains the lowest validation perplexity throughout training.}
  \label{fig:ppl-curves}
\end{figure}

As shown in Figure~\ref{fig:ppl-curves}, the validation perplexity curves remain separated throughout training. The advantage of \texttt{dual\_film} appears early and persists across the training trajectory rather than emerging only at the last evaluation step.

\subsubsection{Multi-Seed Stability}

We further evaluate key variants over three random seeds (42/43/44), as shown in Table~\ref{tab:seed-stability}. In the structural comparison, \texttt{dual\_film} achieves a mean \texttt{final\_val\_ppl} of $5.4117 \pm 0.0079$, outperforming \texttt{dual\_concat} at $5.9822 \pm 0.0194$ and \texttt{single\_global} at $6.1909 \pm 0.0104$. In the coordination comparison, FiLM reaches $5.4579 \pm 0.0169$, outperforming Add at $5.9451 \pm 0.0102$ and Cross at $6.4357 \pm 0.0986$. The ranking is consistent across all three seeds, indicating that the observed gain is not a random initialization artifact.

\begin{table}[t]
  \centering
  \caption{Three-seed stability results on TinyShakespeare.}
  \label{tab:seed-stability}
  \begin{tabular}{lcc}
    \toprule
    Variant & mean final\_val\_ppl & std final\_val\_ppl \\
    \midrule
    dual\_film & 5.4117 & 0.0079 \\
    dual\_concat & 5.9822 & 0.0194 \\
    single\_global & 6.1909 & 0.0104 \\
    \midrule
    FiLM & 5.4579 & 0.0169 \\
    Add & 5.9451 & 0.0102 \\
    Cross & 6.4357 & 0.0986 \\
    \bottomrule
  \end{tabular}
\end{table}

\begin{figure}[t]
  \centering
  \includegraphics[width=0.88\linewidth]{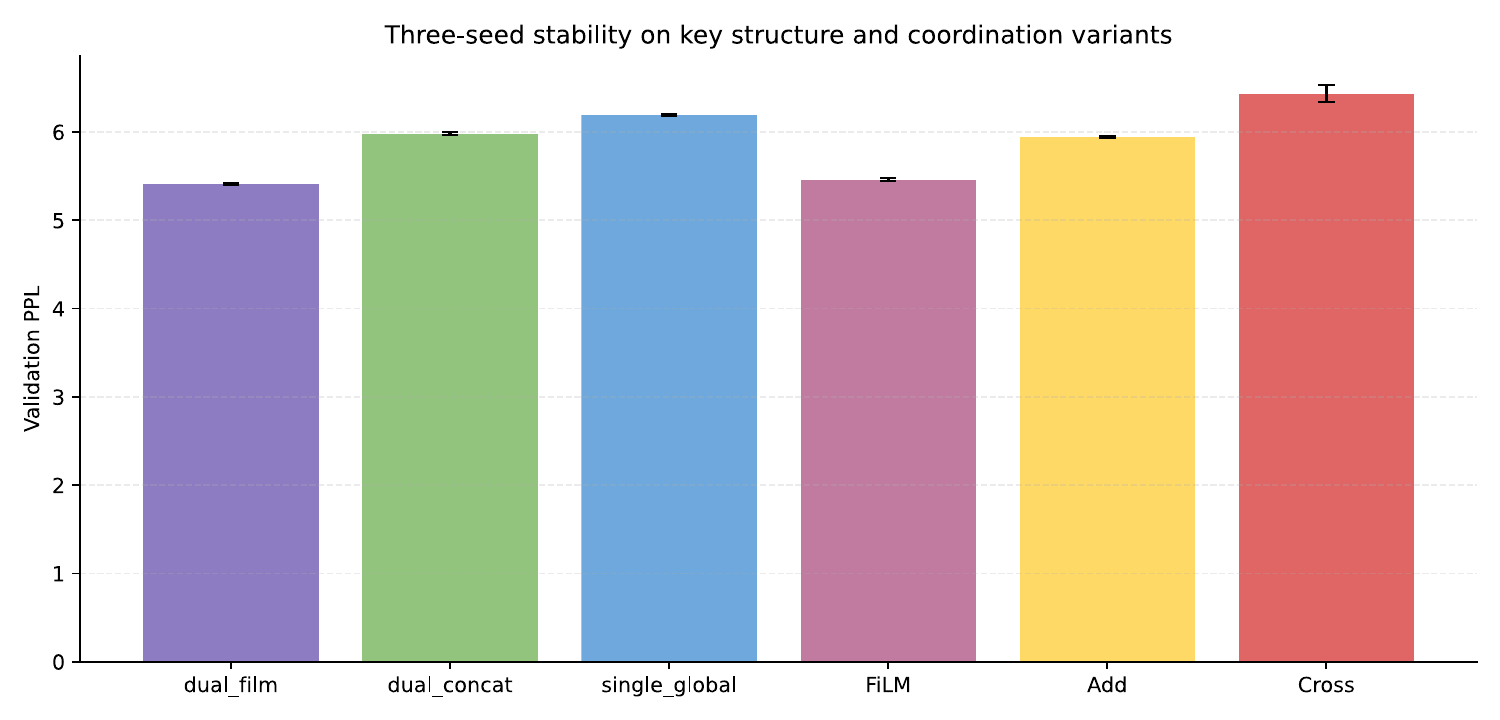}
  \caption{Validation perplexity mean and standard deviation over three random seeds for key structural and coordination variants.}
  \label{fig:seed-stability}
\end{figure}

Figure~\ref{fig:seed-stability} visualizes the same results. Both the key structural variants and the coordination variants show low variance, while \texttt{dual\_film} and FiLM achieve the best mean performance.

\subsubsection{Cross-Dataset Validation}

To reduce the concern that the result holds only on one small corpus, we replicate the key comparisons on a second real-text setting: a 1M-character subset of WikiText-2. The training budget is similar to TinyShakespeare, but the text style is substantially different. Unlike our earlier stage, which only validated the coordination module, we now include the full structural comparison. The results in Table~\ref{tab:wiki-structure} show that \texttt{dual\_film} again performs best, with \texttt{final\_val\_loss}=1.2745 and \texttt{final\_val\_ppl}=3.5768, outperforming \texttt{single\_global} (3.7319), \texttt{single\_local} (3.6938), and \texttt{dual\_concat} (3.7051). This supports the claim that under a fixed lightweight width, the full FiLM-coordinated dual-branch structure remains better than same-width single-branch and weakened dual-branch baselines across different text styles.

As a complementary observation, the coordination-module comparison under the same dual-branch skeleton preserves the same trend: FiLM reaches 5.0810, outperforming Gate at 5.5938, Add at 5.7625, and Cross at 6.2992. In terms of efficiency, \texttt{dual\_film} has latency around 4.96 ms, compared with 1.25/1.48/2.39 ms for \texttt{single\_global}/\texttt{single\_local}/\texttt{dual\_concat}. These results support a careful conclusion: the FiLM structure brings clear accuracy gains, but the current prototype is not primarily a speed-oriented design.

\begin{table}[t]
  \centering
  \caption{Results on the second real-text setting: WikiText-2 1M subset.}
  \label{tab:wiki-structure}
  \begin{tabular}{lccc}
    \toprule
    Structure variant & final\_val\_loss & final\_val\_ppl & latency (ms) \\
    \midrule
    single\_global & 1.3169 & 3.7319 & 1.25 \\
    single\_local & 1.3067 & 3.6938 & 1.48 \\
    dual\_concat & 1.3097 & 3.7051 & 2.39 \\
    dual\_film & 1.2745 & 3.5768 & 4.96 \\
    \bottomrule
  \end{tabular}
\end{table}

\begin{figure}[t]
  \centering
  \includegraphics[width=0.82\linewidth]{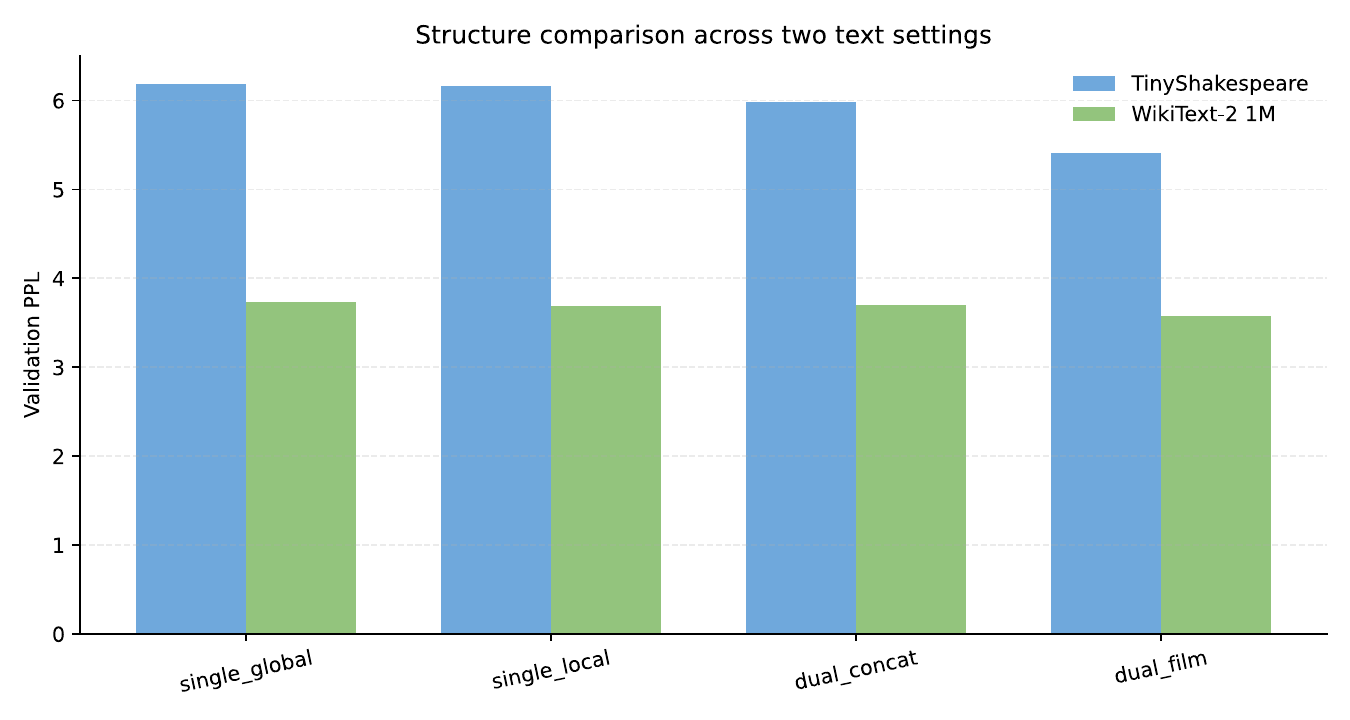}
  \caption{Structural comparison across TinyShakespeare and WikiText-2 1M. \texttt{dual\_film} achieves the best validation perplexity under the fixed lightweight width setting on both datasets.}
  \label{fig:cross-dataset}
\end{figure}

Figure~\ref{fig:cross-dataset} summarizes the structural ranking across the two real-text settings. \texttt{dual\_film} consistently achieves the lowest validation perplexity among the same-width structural variants.

To address the concern that the gain may come only from using more parameters, we further construct parameter-matched single-branch fairness baselines, shown in Table~\ref{tab:fairness}. Specifically, we widen the single-branch models from \texttt{dim=192} to \texttt{dim=360}, resulting in approximately 6.49M parameters for both \texttt{single\_global} and \texttt{single\_local}, close to the 6.36M parameters of \texttt{dual\_film}. Under this stricter comparison, \texttt{single\_global} and \texttt{single\_local} achieve \texttt{final\_val\_ppl}=3.5197 and 3.5226, slightly better than the 3.5768 of \texttt{dual\_film}. This indicates that the structural gain is clearly supported under a fixed lightweight width, but is not yet preserved under a strict parameter-matched perspective. A more accurate conclusion is therefore that the proposed structure improves over weakened structures and small same-width baselines, while its parameter efficiency remains an open optimization target.

\begin{table}[t]
  \centering
  \caption{Parameter-matched fairness comparison on the WikiText-2 1M subset.}
  \label{tab:fairness}
  \begin{tabular}{lcccc}
    \toprule
    Variant & Parameters & final\_val\_loss & final\_val\_ppl & latency (ms) \\
    \midrule
    dual\_film (dim=192) & 6.36M & 1.2745 & 3.5768 & 4.96 \\
    single\_global (dim=360) & 6.49M & 1.2584 & 3.5197 & 1.27 \\
    single\_local (dim=360) & 6.49M & 1.2592 & 3.5226 & 3.54 \\
    \bottomrule
  \end{tabular}
\end{table}

\begin{figure}[t]
  \centering
  \includegraphics[width=0.92\linewidth]{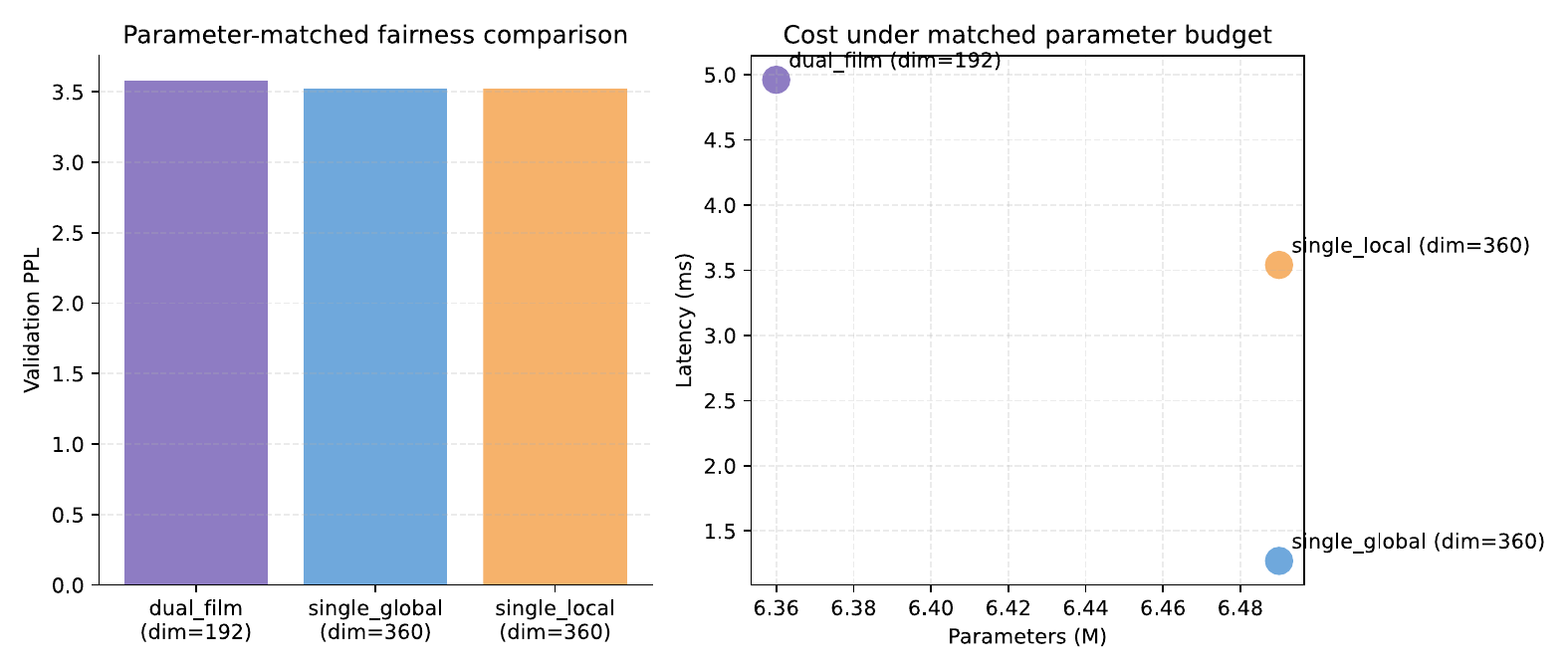}
  \caption{Parameter-matched fairness comparison. The current \texttt{dual\_film} model is competitive under a fixed lightweight width, but widened single-branch baselines perform slightly better under near-matched parameter counts.}
  \label{fig:fairness}
\end{figure}

Figure~\ref{fig:fairness} visualizes this accuracy-cost relation more clearly. The current \texttt{dual\_film} structure has a real structural advantage under the fixed lightweight regime, but widened single-branch baselines still achieve slightly lower perplexity under near-matched parameter budgets.

\subsubsection{Positioning of the Coordination Comparison}

It is important to emphasize that the coordination comparison is not the main claim of this paper. Its purpose is to answer a narrower question: once a dual-branch structure is adopted, what type of coordination best matches that structure? The current evidence supports viewing FiLM as an internal structural component of the new architecture rather than reducing the paper to a generic fusion benchmark.

\subsection{Structural Ablations}

\begin{table}[t]
  \centering
  \caption{Structural ablation results.}
  \label{tab:ablations}
  \resizebox{\linewidth}{!}{%
  \begin{tabular}{llccc}
    \toprule
    Ablation axis & Variant & final\_val\_loss & final\_val\_ppl & Extra metric \\
    \midrule
    Modulation direction & film\_bidirectional & 1.6911 & 5.4209 & -- \\
     & film\_local\_to\_global & 1.7081 & 5.5237 & -- \\
     & film\_global\_to\_local & 1.7260 & 5.6231 & -- \\
    \midrule
    Local window size & window\_32 & 1.7103 & 5.5360 & latency $\sim 5.04$ ms \\
     & window\_64 & 1.6930 & 5.4312 & latency $\sim 5.11$ ms \\
     & window\_128 & 1.6992 & 5.4647 & latency $\sim 4.97$ ms \\
     & window\_256 & 1.6881 & 5.4039 & latency $\sim 5.08$ ms \\
    \midrule
    FiLM generator capacity & hidden\_ratio\_0.5 & 1.6972 & 5.4535 & peak\_memory 52.51 MB \\
     & hidden\_ratio\_1.0 & 1.6950 & 5.4415 & peak\_memory 55.34 MB \\
     & hidden\_ratio\_2.0 & 1.6911 & 5.4209 & peak\_memory 61.17 MB \\
     & hidden\_ratio\_4.0 & 1.6852 & 5.3878 & peak\_memory 73.66 MB \\
    \bottomrule
  \end{tabular}
  }
\end{table}

\subsubsection{Modulation Direction}

Directional ablation tests whether mutual modulation is necessary. Bidirectional FiLM outperforms both one-way settings: \texttt{film\_bidirectional} reaches \texttt{final\_val\_loss}=1.6911, compared with 1.7081 for \texttt{film\_local\_to\_global} and 1.7260 for \texttt{film\_global\_to\_local}. This suggests that one branch merely patching the other is not sufficient; two-way coordination better matches the structural goal of jointly modeling global and local dependencies.

\subsubsection{Local Window Size}

Window-size ablation shows that the structure is not highly sensitive to the local receptive field. \texttt{window\_64} and \texttt{window\_256} are both close to the best setting, with \texttt{final\_val\_loss} around 1.6930 and 1.6881, while \texttt{window\_32} is slightly weaker at 1.7103. The latency differences are small, with \texttt{window\_64-256} all lying around 4.97--5.11 ms. This indicates that the proposed structure does not depend on a narrow window-size sweet spot.

\subsubsection{FiLM Generator Capacity}

The FiLM-generator capacity ablation shows that \texttt{hidden\_ratio\_0.5}, \texttt{2.0}, and \texttt{4.0} all work stably. The best result comes from \texttt{hidden\_ratio\_4.0} with \texttt{final\_val\_loss}=1.6852, but the smaller settings already remain close. In terms of cost, \texttt{hidden\_ratio\_4.0} increases peak memory to about 73.66 MB, compared with 52.51/55.34 MB for \texttt{0.5/1.0}. This suggests that the gain does not merely come from enlarging the modulation MLP.

\subsection{Cost and Mechanistic Analysis}

\begin{table}[t]
  \centering
  \caption{Inference latency comparison across sequence lengths 128/256/512/1024 (ms per decoding step).}
  \label{tab:efficiency}
  \begin{tabular}{lcccc}
    \toprule
    Variant & seq\_len=128 & seq\_len=256 & seq\_len=512 & seq\_len=1024 \\
    \midrule
    Add & 1.22 & 2.03 & 4.21 & 8.33 \\
    Concat & 1.31 & 2.17 & 4.48 & 8.56 \\
    FiLM & 2.52 & 3.97 & 7.35 & 13.51 \\
    Cross & 2.68 & 4.85 & 9.76 & 16.25 \\
    \bottomrule
  \end{tabular}
\end{table}

\subsubsection{Efficiency and Cost}

The efficiency study shows that FiLM is not the fastest coordination option in the current prototype. In long-sequence benchmarking, Add and Concat remain the lightest baselines, while Cross has the steepest latency curve. FiLM performs better in accuracy, but does not exhibit a universal speed advantage over lightweight alternatives. A more careful conclusion is therefore that FiLM provides stronger structural gains and better interpretability, while its efficiency advantage is mainly relative to Cross rather than absolute.

\begin{figure}[t]
  \centering
  \includegraphics[width=0.78\linewidth]{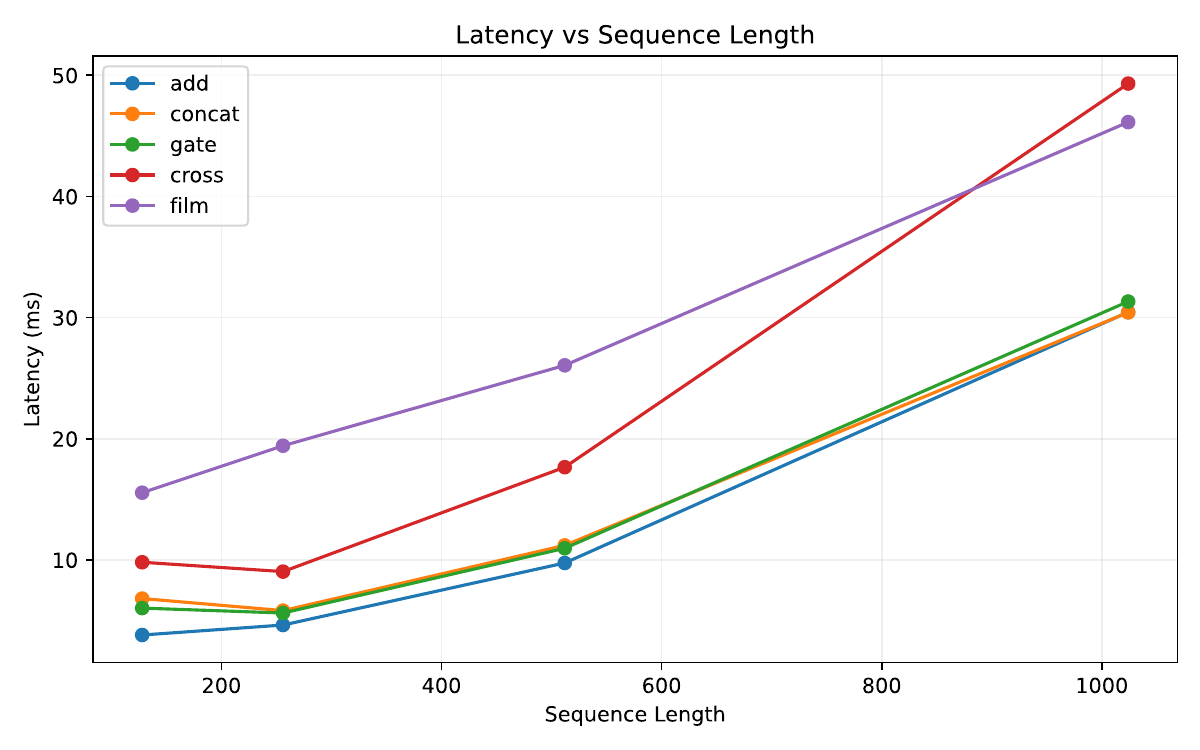}
  \caption{Latency versus sequence length for different coordination methods. Cross scales worst, while FiLM lies between lightweight fusion and heavy interaction.}
  \label{fig:latency}
\end{figure}

Figure~\ref{fig:latency} shows the latency trend as sequence length grows. Cross scales worst, while FiLM lies between lightweight fusion and heavier interaction.

\subsubsection{Modulation Statistics and Input Response}

Mechanistic analysis further shows that the learned FiLM behavior is not uniform scaling. As summarized in Table~\ref{tab:film-stats}, the training-time \texttt{topk\_gamma\_ratio} is around 2.74--2.80, suggesting that strong modulation is concentrated on a subset of active channels. Deeper layers exhibit stronger modulation than shallow layers, with \texttt{mean\_abs\_gamma} around 0.65 at layer 4 versus 0.52 at layer 1. Different input types also trigger different modulation strengths, with \texttt{code} and \texttt{long\_dependency} eliciting stronger responses. This supports the interpretation that FiLM learns input-dependent structural calibration.

\begin{table}[t]
  \centering
  \caption{Mechanistic statistics of learned FiLM modulation.}
  \label{tab:film-stats}
  \begin{tabular}{lc}
    \toprule
    Statistic & Value \\
    \midrule
    training-time topk\_gamma\_ratio & 2.74--2.80 \\
    mean\_abs\_gamma on repetition inputs & 0.6065 \\
    mean\_abs\_gamma on long\_dependency inputs & 0.6184 \\
    mean\_abs\_gamma on narrative inputs & 0.6195 \\
    mean\_abs\_gamma on code inputs & 0.6529 \\
    layer-1 mean\_abs\_gamma & $\sim 0.52$ \\
    layer-4 mean\_abs\_gamma & $\sim 0.65$ \\
    \bottomrule
  \end{tabular}
\end{table}

\begin{figure}[t]
  \centering
  \includegraphics[width=0.7\linewidth]{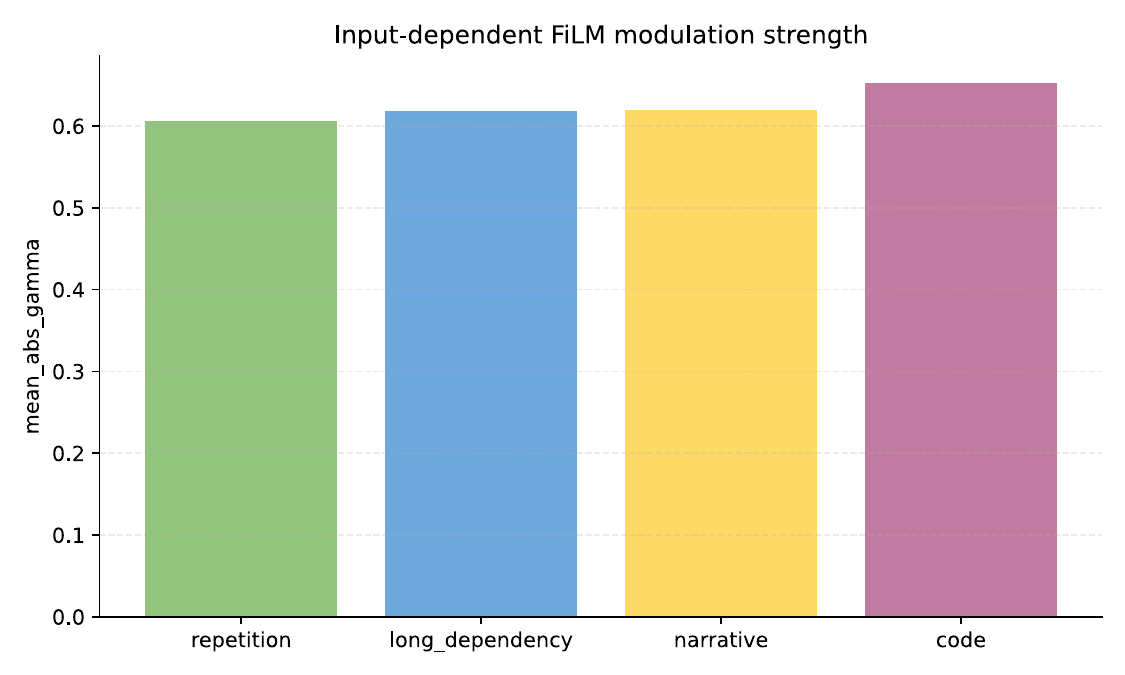}
  \caption{Average FiLM modulation strength across input categories. Code-like and long-dependency inputs induce stronger modulation.}
  \label{fig:adaptivity}
\end{figure}

Figure~\ref{fig:adaptivity} visualizes modulation strength across different input types. Code-like and long-dependency inputs consistently induce stronger modulation than simpler repetition-style inputs.

\subsection{Discussion and Limitations}

Although the current evidence is sufficient for a structural prototype paper, the limitations are clear. First, all experiments are conducted at a small language-modeling scale and cannot yet be directly extrapolated to large-scale LLMs. Second, the cross-dataset validation, while stronger than before, still covers only one additional real-text subset under a controlled budget. Third, the parameter-matched fairness comparison shows that widened single-branch baselines can slightly surpass the current \texttt{dual\_film} model on WikiText-2 1M, indicating that the structural gain is clearest under fixed lightweight widths rather than under strict parameter matching. Finally, the present FiLM implementation does not provide a universal speed advantage over all lightweight baselines. These observations suggest that future work should focus on longer contexts, stronger baselines, and especially improved parameter efficiency.

\subsection{Section Summary}

Overall, the available evidence supports the core structural claim of this paper: under a fixed lightweight backbone width, the dual-branch design is useful, and FiLM serves as an effective intra-layer coordination mechanism that stabilizes and strengthens this structural gain. At the same time, the fairness comparison sets an important boundary on the claim: the current method is best positioned as a small-scale but closed-loop structural prototype rather than a fully dominant architecture under strict budget matching.

\section{Conclusion}

We proposed a FiLM-coordinated dual-branch Transformer for jointly modeling global and local dependencies within the same layer. Rather than framing the study as a search for the best generic fusion head, we treated FiLM as a structural coordination mechanism inside a dual-branch Transformer and evaluated the architecture from that perspective. The experiments show that under a fixed lightweight configuration, the full dual-branch FiLM structure outperforms single-branch baselines and weakened dual-branch variants. Multi-seed and cross-dataset results support the stability of this effect, while mechanistic analysis shows that FiLM learns input-dependent, layer-dependent, and channel-selective modulation rather than static scaling.

The main value of this work is therefore not a new fusion operator, but a clearer formulation of intra-layer global-local coordination as a structural design problem in Transformers, together with an interpretable and empirically supported solution. The present study remains a small-scale prototype and should be extended to larger models, longer contexts, stricter budget matching, and stronger baselines. Nevertheless, the current results suggest that FiLM-coordinated dual-branch structures are a promising direction for further investigation.

\end{document}